\documentclass[11pt,a4paper]{article}
\usepackage[hyperref]{naaclhlt2018}
\usepackage{times}
\usepackage{latexsym}

\usepackage{caption}
\usepackage{subcaption}
\usepackage{amsmath}
\usepackage{multirow}
\usepackage{booktabs}
\usepackage{forest}
\forestset{
    nice empty nodes/.style={
        for tree={
            s sep=0.1em, 
            l sep=0.33em,
            inner ysep=0.4em, 
            inner xsep=0.05em,
            l=0,
            calign=midpoint,
            fit=tight,
            where n children=0{
               tier=word,
               minimum height=1.25em,
            }{},
            where n children=2{
               l-=1em,
            }{},
            parent anchor=south,
            child anchor=north,
            delay={if content={}{
                    inner sep=0pt,
                    edge path={\noexpand\path [\forestoption{edge}] 
                    			(!u.parent anchor) 
                               -- (.south)\forestoption{edge label};}
                }{}}
        },
    },
}

\DeclareMathOperator{\MAX}{\textsc{max}}
\DeclareMathOperator{\MIN}{\textsc{min}}
\DeclareMathOperator{\MED}{\textsc{med}}
\DeclareMathOperator{\SM}{\textsc{sm}}

\usepackage{url}
\usepackage{scalefnt}
\usepackage{graphicx}
\usepackage{tikz}
\usepackage{todonotes}
\makeatletter
\newcommand{\@emptybiblabel}[1]{}
\makeatother

\aclfinalcopy % Uncomment this line for the final submission
\def\aclpaperid{3052} %  Enter the acl Paper ID here

\title{ListOps: A Diagnostic Dataset for Latent Tree Learning}

\author{
Nikita Nangia$^{1}$\\
\texttt{\small nikitanangia@nyu.edu}
\And
Samuel R.~Bowman$^{1,2,3}$\\
\texttt{\small bowman@nyu.edu}
\AND
$^{1}$\normalfont Center for Data Science\\New York University\\60 Fifth Avenue\\New York, NY 10011\And
$^{2}$\normalfont Dept. of Linguistics\\New York University\\10 Washington Place\\New York, NY 10003\And
$^{3}$\normalfont Dept. of Computer Science\\New York University\\60 Fifth Avenue\\New York, NY 10011
} 

\date{}

\begin{document}
\maketitle
\begin{abstract}
Latent tree learning models learn to parse a sentence without syntactic supervision, and use that parse to build the sentence representation. Existing work on such models has shown that, while they perform well on tasks like sentence classification, they do not learn grammars that conform to any plausible semantic or syntactic formalism \citep{williams2018learning}. Studying the parsing ability of such models in natural language can be challenging due to the inherent complexities of natural language, like having several valid parses for a single sentence. In this paper we introduce ListOps, a toy dataset created to study the parsing ability of latent tree models. ListOps sequences are in the style of prefix arithmetic. The dataset is designed to have a single correct parsing strategy that a system needs to learn to succeed at the task. We show that the current leading latent tree models are unable to learn to parse and succeed at ListOps. These models achieve accuracies worse than purely sequential RNNs.
\end{abstract}

%%%%%%%%%%%% INTRODUCTION %%%%%%%%%%%%
\section{Introduction}
Recent work on \textit{latent tree learning} models \citep{yogatama2016learning,maillard2017jointly,choi2017unsupervised,williams2018learning} has introduced new methods of training tree-structured recurrent neural networks \citep[TreeRNNs;][]{socher2011semi} without ground-truth parses. These latent tree models learn to parse with indirect supervision from a downstream semantic task, like sentence classification. They have been shown to perform well at sentence understanding tasks, like textual entailment and sentiment analysis, and they generally outperform their TreeRNN counterparts that use parses from conventional parsers.

Latent tree learning models lack direct syntactic supervision, so they are not being pushed to conform to expert-designed grammars, like the Penn Treebank \citep[PTB;][]{ptb}. Theoretically then, they have the freedom to learn whichever grammar is best suited for the task at hand. However, \citet{williams2018learning} show that current latent tree learning models do not learn grammars that follow recognizable semantic or syntactic principles when trained on natural language inference. Additionally, the learned grammars are not consistent across random restarts. This begs the question, do these models fail to learn useful grammars because it is unnecessary for the task? Or do they fail because they are incapable of learning to parse? In this paper we introduce the ListOps datasets which is designed to address this second question.

\begin{figure}[t]
\centering
\scalebox{1.2}{
\begin{forest}
 shape=coordinate,
 where n children=0{
   tier=word
 }{},
 nice empty nodes
[ [ [ [ [ [\textsc{[max}] [2] ] [9] ] [ [ [ [\textsc{[min}] [4] ] [7] ] [\textsc{]}] ] ] [0] ] [\textsc{]}] ]
\end{forest}
}
\caption{\label{fig:exampletrees} Example of a parsed ListOps sequence. The parse is left-branching within each list, and each constituent is either a partial list, an integer, or the final closing bracket.} 
\end{figure}

Since natural language is complex, there are often multiple valid parses for a single sentence. Furthermore, as was shown in \citet{williams2018learning}, using sensible grammars is not necessary to do well at some existing natural language datasets. Since our primary objective is to study a system's ability to learn a correct parsing strategy, we build a toy dataset, ListOps, that primarily tests a system's parsing ability. ListOps is in the style of prefix arithmetic; it is comprised of deeply nested lists of mathematical operations and a list of single-digit integers.

\begin{figure*}[!ht]

	\centering
	\scalebox{1.0}{
	\begin{forest}
		shape=coordinate,
		where n children=0{
			tier=word
		}{},
		nice empty nodes
		[ [ [ [ [ [ [ [ [ [ [ [ [\textsc{[max}] [ [\textsc{[med}] [ [\textsc{[med}] [1] ] ] ] [\textsc{[sm}] ] [3] ] [1] ] [3] ] [\textsc{]}] ] [9] ] [\textsc{]}] ] [6] ] [\textsc{]}] ] [5] ] [\textsc{]}] ]
	\end{forest}}
	\hspace{2.0em}
	\scalebox{1.0}{
	\begin{forest}
		shape=coordinate,
		where n children=0{
			tier=word
		}{},
		nice empty nodes
		[ [ [\textsc{[max}] [ [\textsc{[med}] [ [\textsc{[med}] [1] ] ] ] [ [\textsc{[sm}] [ [ [ [ [ [3] [ [1] [3] ] ] [\textsc{]}] ] [9] ] [ [\textsc{]}] [ [ [6] [\textsc{]}] ] [5] ] ] ] [\textsc{]}] ] ] ]
\end{forest}}
	\vspace{0.3em}\\
	\hspace{1.5em} \textit{Truth: 6; Pred: 5}
	\hspace{13.5em} \textit{Truth: 6; Pred: 5}
	
	\vspace{2.5em}

	\scalebox{1.0}{
	\begin{forest}
		shape=coordinate,
		where n children=0{
			tier=word
		}{},
		nice empty nodes
		[ [ [ [ [ [\textsc{[sm}] [\textsc{[sm}] ] [\textsc{[sm}] ] [ [ [\textsc{[max}] [ [5] [ [6] [\textsc{]}] ] ] ] [ [2] [\textsc{]}] ] ] ] [ [0] [ [\textsc{]}] [ [5] [ [0] [8] ] ] ] ] ] [ [6] [\textsc{]}] ] ]
	\end{forest}}\hspace{2.0em}
	\scalebox{1.0}{
	\begin{forest}
		shape=coordinate,
		where n children=0{
			tier=word
		}{},
		nice empty nodes
		[ [\textsc{[sm}] [ [\textsc{[sm}] [ [\textsc{[sm}] [ [ [ [ [ [ [ [\textsc{[max}] [5] ] [6] ] [\textsc{]}] ] [2] ] [\textsc{]}] ] [ [ [0] [\textsc{]}] ] [ [5] [0] ] ] ] [ [ [8] [6] ] [\textsc{]}] ] ] ] ] ]
\end{forest}}
	\vspace{0.3em}\\
	\hspace{2.4em} \textit{Truth: 7; Pred: 7}
	\hspace{13.5em} \textit{Truth: 7; Pred: 2}

	\vspace{2.5em}

	\scalebox{1.0}{
	\begin{forest}
		shape=coordinate,
		where n children=0{
			tier=word
		}{},
		nice empty nodes
		[ [ [ [ [ [ [\textsc{[med}] [6] ] [ [ [ [ [\textsc{[med}] [3] ] [2] ] [2] ] [\textsc{]}] ] ] [~8~] ] [5] ] [ [ [ [ [\textsc{[med}] [8] ] [6] ] [2] ] [\textsc{]}] ] ] [\textsc{~~~]~~~~}] ]
	\end{forest}}\hspace{2.0em}
	\scalebox{1.0}{
	\begin{forest}
		shape=coordinate,
		where n children=0{
			tier=word
		}{},
		nice empty nodes
		[ [ [ [\textsc{[med}] [6] ] [ [ [\textsc{[med}] [ [3] [ [2] [2] ] ] ] [\textsc{]}] ] ] [ [ [ [8] [5] ] [ [ [\textsc{[med}] [ [8] [ [6] [2] ] ] ] [\textsc{]}] ] ] [\textsc{]}] ] ]
\end{forest}}
	\vspace{0.3em}\\
	\hspace{2.8em} \textit{Truth: 6; Pred: 6}
	\hspace{13.6em} \textit{Truth: 6; Pred: 5}
	
\caption{\label{fig:trees}\textit{Left:} Parses from RL-SPINN model. \textit{Right:} Parses from ST-Gumbel model. For the first set of examples in the top row, both each models predict the wrong value \textit{(truth:~6, pred:~5)}. In the second row, RL-SPINN predicts the correct value \textit{(truth:~7)} while ST-Gumbel does not \textit{(pred:~2)}. In the third row, RL-SPINN predicts the correct value \textit{(truth:~6)} and generates the same parse as the ground-truth tree; ST-Gumbel predicts the wrong value \textit{(pred:~5)}.}

\end{figure*}
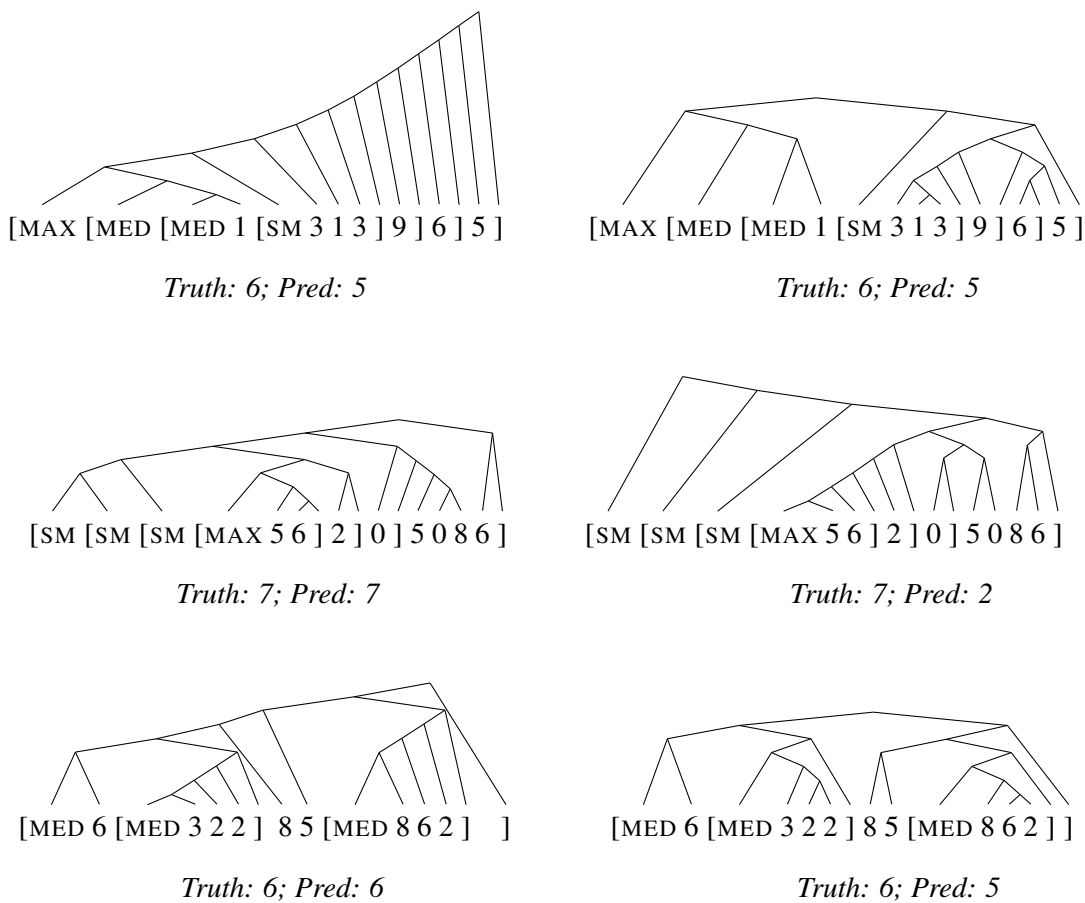

The ListOps sequences are generated with a reference parse, and this parse corresponds to the simplest available strategy for interpretation. We are unaware of reasonably effective strategies that differ dramatically from our reference parses. If a system is given the ground-truth parses, it is trivially easy to succeed at the task. However, if the system does not have the reference parses, or is unable to learn to parse, doing well on ListOps becomes dramatically more difficult. Therefore, we can use ListOps as a litmus test and diagnostic tool for studying latent tree learning models. ListOps is an environment where parsing is essential to success. So if a latent tree model is able to achieve high accuracy in this rigid environment, it indicates that the model is able to learn a sensible parsing strategy. Conversely, if it fails on ListOps, it may suggest that the model is simply incapable of learning to parse.

%%%%%%%%%%%%%% RELATED WORK %%%%%%%%%%%%%%%%
\section{Related Work}\label{sec:related}

To the best of our knowledge, all existing work on latent tree models studies them in a natural language setting.  \citet{williams2018learning} experiment with two leading latent tree models on the textual entailment task, using the SNLI \citep{snli:emnlp2015} and MultiNLI corpora \citep{williams2018broad}. The \citet{williams2018learning} analysis studies the models proposed by \citet{yogatama2016learning} (which they call RL-SPINN) and \citet{choi2017unsupervised} (which they call ST-Gumbel). A third latent tree learning model, which is closely related to ST-Gumbel, is presented by \citet{maillard2017jointly}. 

All three models make use of TreeLSTMs \citep{tai2015improved} and learn to parse with distant supervision from a downstream semantic objective. The RL-SPINN model uses the REINFORCE algorithm \citep{williams1992simple} to train the model's parser. The parser makes discrete decisions and cannot be trained with backpropogation. 

The model \citet{maillard2017jointly} present uses a CYK-style \citep{cocke1969, younger1967, kasami1965} chart parser to compute a soft combination of all valid binary parse trees. This model computes $O(N^2)$ possible tree nodes for $N$ words, making it computationally intensive, particularly on ListOps which has very long sequences.

The ST-Gumbel model uses a similar data structure to \citeauthor{maillard2017jointly}, but instead utilizes the Straight-Through Gumbel-Softmax estimator \citep{jang2016categorical} to make discrete decisions in the forward pass and select a single binary parse.

Our work, while on latent tree learning models, is with a toy dataset designed to study parsing ability. There has been some previous work on the use of toy datasets to closely study the performance of systems on natural language processing tasks. For instance, \citet{weston2015towards} present bAbI, a set of toy tasks for to testing Question-Answering systems. The tasks are designed to be prerequisites for any system that aims to succeed at language understanding. The bAbI tasks have influenced the development of new learning algorithms \citep{sukhbaatar15end, Kumar2016AskMA, Peng2015TowardsNN}.

%%%%%%%%%%%%%% DATASET %%%%%%%%%%%%%%%%
\section{Dataset}\label{sec:dataset}

\paragraph{Description} The ListOps examples are comprised of summary operations on lists of single-digit integers, written in prefix notation. The full sequence has a corresponding solution which is also a single-digit integer, thus making it a ten-way balanced classification problem. For example, $ [\MAX\ 2\ 9\ [\MIN 4\ 7\ ]\ 0\ ] $ has the solution 9. Each operation has a corresponding closing square bracket that defines the list of numbers for the operation. In this example, $\MIN$ operates on $\{4, 7\}$, while $\MAX$ operates on $\{2, 9, 4, 0\}$. The correct parse for this example is shown in Figure~\ref{fig:exampletrees}. As with this example, the reference parses in ListOps are left-branching within each list. If they were right-branching, the model would always have to maintain the entire list in memory. This is because the summary statistic for each list is dependent on the type of operation, and the operation token appears first in prefix notation. 

Furthermore, we select a small and easy operation space to lower output set difficulty. The operations that appear in ListOps are:

\begin{itemize}
	\item \textsc{max}: the largest value of the given list. For the list $\{8, 12, 6, 3\}$, 12 is the $\MAX$.	\item \textsc{min}: the smallest value of the given list. For the list $\{8, 12, 6, 3\}$, 3 is the $\MIN$.
	\item \textsc{med}: the median value of the given list. For the list $\{8, 12, 6, 3\}$, 7 is the $\MED$.
	\item \textsc{sum\_mod} (\textsc{sm}): the sum of the items in the list, constrained to a single digit by the use of the modulo-10 operator. For the list $\{8, 12, 6, 3\}$, 9 is the $\SM$.
\end{itemize}

ListOps is constructed such that it is trivially easy to solve  if a model has access to the ground-truth parses. However, if a model does not have the parses, or is unable to learn to parse correctly, it may have to maintain a large stack of information to arrive at the correct solution. This is particularly true as the sequences become long and have many nested lists.

\paragraph{Efficacy} We take an empirical approach to determine the efficacy of the ListOps dataset to test parsing capability. ListOps should be trivial to solve if a model is given the ground-truth parses. Therefore, a tree-structured model that is provided with the parses should be able to achieve near 100\% accuracy on the task. So, to establish the upper-bound and solvability of the dataset, we use a TreeLSTM as one of our baselines.

Conversely, if the ListOps dataset is adequately difficult, then a strong sequential model should not perform well on the dataset. We use an LSTM \citep{hochreiter1997long} as our sequential baseline. 

We run extensive experiments on the ListOps dataset to ensure that the TreeLSTM does consistently succeed while the LSTM fails. We tune the model size, learning rate, L2 regularization, and decay of learning rate (the learning rate is lowered at every epoch when there has been no gain). We require that the TreeLSTM model does well at a relatively low model size. We further ensure that the LSTM, at an order of magnitude greater model size, is still unable to solve ListOps. Therefore, we build the dataset and establish its effectiveness as a diagnostic task by maximizing this RNN--TreeRNN gap.

Theoretically, this RNN--TreeRNN gap arises because an RNN of fixed size does not have the capacity to store all the necessary information. More concretely, we know that each of the operations in ListOps can be computed by passing over the list of integers with a constant amount of memory. For example, to compute the \textsc{max}, the system only needs to remember the largest number it has seen in the operation's list. 
As an RNN reads a sequence, if it is in the middle of the sequence, it will have read many operations without closed parentheses, i.e. without terminating the lists. Therefore, it has to maintain the state of all the open operations it has read. So the amount of information the RNN has to maintain grows linearly with tree depth. As a result, once the trees are deep enough, an RNN with a fixed-size memory cannot effectively store and retrieve all the necessary information. % "the necessary" or just "necessary"?

For a TreeRNN, every constituent in ListOps is either a partial list, an integer, or the final closing bracket. For example, in Figure~\ref{fig:exampletrees}, the first constituent, $( [\MAX, 2, 9 )$, is a partial list. So, the amount of information the TreeLSTM has to store at any given node is no greater than the small amount needed to process one list. Unlike with an RNN, this small amount of information at each node does not grow with tree depth. Consequently, TreeRNNs can achieve high accuracy at ListOps with very low model size, while RNNs require higher capacity to do well.

\begin{figure}[t!]
  \centering
    \includegraphics[width=0.5\textwidth]{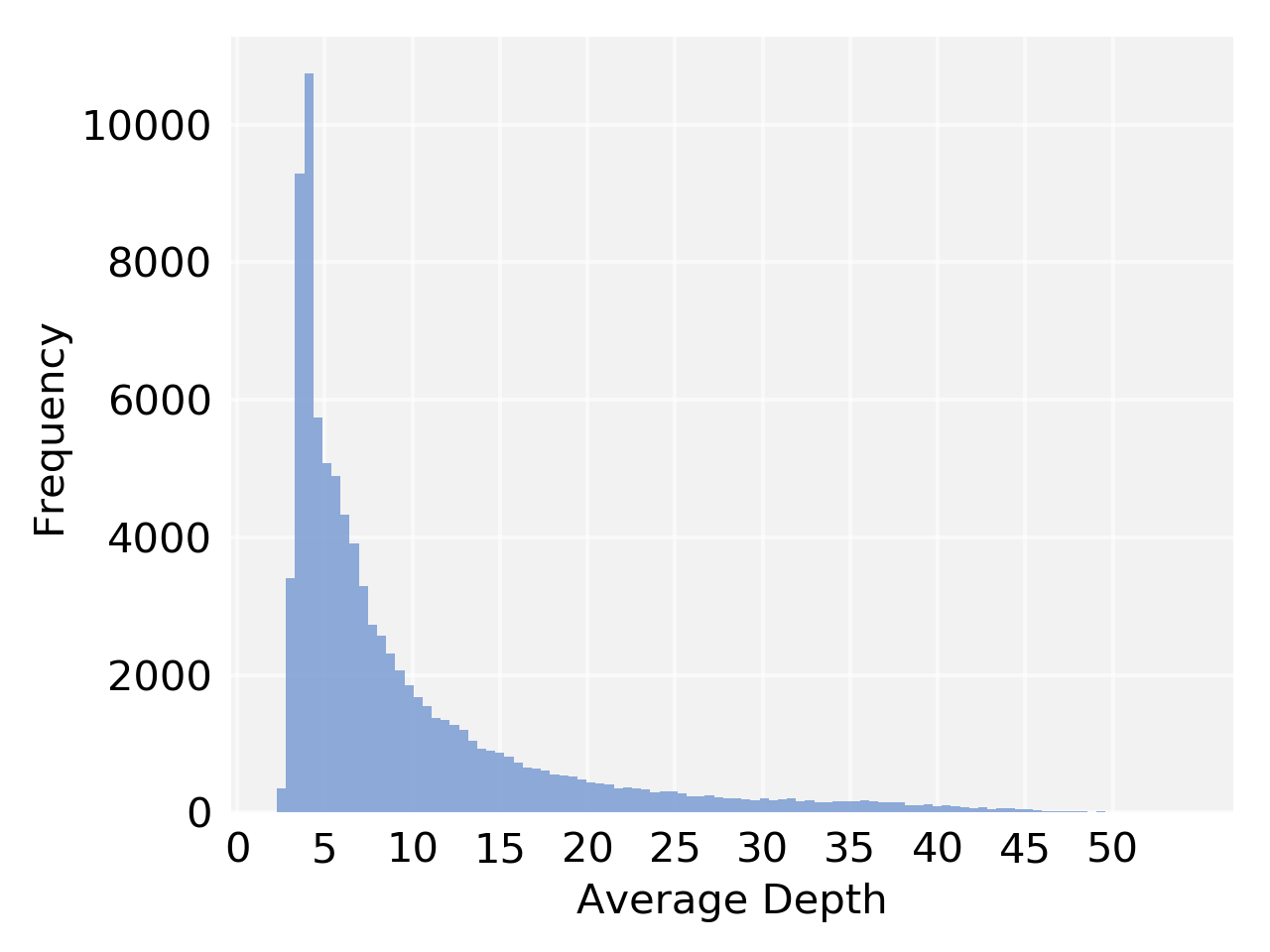}
   	\caption{Distribution of average tree depth in the ListOps training dataset.}
	\label{fig:depths}
\end{figure}

\paragraph{Generation} The two primary variables that determine the difficulty of the ListOps dataset are tree depth and the function space of mathematical operations. We found tree depth to be an essential variable in stressing model performance, and in maximizing the RNN--TreeRNN gap. While creating ListOps, we clearly observe that with increasing recursion in the dataset the performance of sequential models falls. Figure~\ref{fig:depths} shows the distribution of tree depths in the ListOps dataset; the average tree depth is 9.6.

As discussed previously, since we are concerned with a model's ability to learn to parse, and not its ability to approximate mathematical operations, we choose a minimal number of operations (\textsc{max, min, med, sm}). In our explorations, we find that these easy-to-compute operations yield bigger RNN--TreeRNN gaps than operations like multiplication.

The ListOps dataset used in this paper has 90k training examples and 10k test examples. During data generation, the operations are selected at random, and their frequency is balanced in the final dataset. We wrote a simple Python script to generate the ListOps data. Variables such as maximum tree-depth, as well as number and kind of operations, can be changed to generate variations on ListOps. One might want to increase the average tree depth if a model with much larger hidden states is being tested. With a very large model size, an RNN, in principle, can succeed at the ListOps dataset presented in this paper. The dataset and data generation script are available on GitHub.\footnote[1]{\label{git}\url{https://github.com/NYU-MLL/spinn/tree/listops-release}}

\section{Models}
We use an LSTM for our sequential baseline, and a TreeLSTM for our tree-structured baseline.
For the latent tree learning models, we use two leading models discussed in Section~\ref{sec:related}: RL-SPINN \citep{yogatama2016learning} and ST-Gumbel \citep{choi2017unsupervised}. We are borrowing the model names from \citet{williams2018learning}.

\paragraph{Training details} All models are implemented in a shared codebase in PyTorch 0.3, and the code is available on GitHub.\textsuperscript{\ref{git}} We do extensive hyperparameter tuning for all baselines and latent tree models. We tune the learning rate, L2 regularization, and rate of learning rate decay. We tune the model size for the baselines in our endeavor to establish the RNN--TreeRNN gap, wanting to ensure that the TreeLSTM, with reference parses, can solve ListOps at a low hidden dimension size, while the LSTM can not solve the dataset at significantly larger hidden sizes.
We test model sizes from 32D to 1024D for the baselines. The model size for latent tree models is tuned to a lesser extent, since a model with parsing ability should have adequate representational power at lower dimensions. We choose the narrower range of model sizes based on how well the TreeLSTM baseline performs at those sizes. We consider latent tree model sizes from 32D to 256D. Note that the latent tree models we train with sizes greater than 128D do not show significant improvement in performance accuracy. 

%We randomly initialize all word-level representations, i.e. representations for the operations, integers, and parenthesis. These representations are updated with the rest of the model. 
For all models, we pass the representation through a 2-layer MLP, followed by a ten-way softmax classifier. We use the Adam optimizer \citep{kingma2014adam} with default values for the beta and epsilon parameters.

%%%%%%%%%%%% RESULTS %%%%%%%%%%%%
\begin{table}[t!]
\small \centering
\begin{tabular}{lcc}
\toprule
\bf Model & \bf ListOps & \bf SNLI  \\
\midrule
\multicolumn{3}{c}{Prior Work: Baselines} \\
\midrule
100D LSTM (Yogatama) & -- & 80.2 \\
300D BiLSTM (Williams) & -- & 81.5 \\
300D TreeLSTM (Bowman) & -- &  80.9 \\
\midrule
\multicolumn{3}{c}{Prior Work: Latent Tree Learning} \\
\midrule
300D RL-SPINN (Williams) & --  & 83.3 \\
300D ST-Gumbel (Choi) & -- & \bf 84.6 \\
100D Soft-Gating (Maillard) & -- &  81.6 \\
\midrule
\multicolumn{3}{c}{This Work: Baselines} \\
\midrule
128D~~ LSTM & 73.3 & --\\
1024D LSTM &  74.4  & --\\
48D~~~~ TreeLSTM & 94.7  & --\\
128D~~ TreeLSTM & \bf 98.7 & --\\
\midrule
\multicolumn{3}{c}{This Work: Latent Tree Learning} \\
\midrule
48D~~ RL-SPINN & 62.3 & -- \\
128D RL-SPINN & 64.8 & --\\
48D~~ ST-Gumbel & 58.5 & --\\
128D ST-Gumbel & 61.0 & --\\
%32D~~ Soft-Gating & 56.0 & --\\
\bottomrule
\end{tabular}  
\caption{\label{tab:acc} \textit{SNLI} shows test set results of models on the Stanford Natural Language Inference Corpus, a sentence classification task. We see that the latent tree learning models outperform the supervised TreeLSTM model. However, on \textit{ListOps}, RL-SPINN and ST-Gumbel have worse performance accuracy than the LSTM baseline.}
\end{table}

\section{ListOps Results}
\paragraph{Baseline models} The results for the LSTM and TreeLSTM baseline models are shown in Table~\ref{tab:acc}. We clearly see the RNN--TreeRNN gap. The TreeLSTM model does well on ListOps at embedding dimensions as low as 48D, while the LSTM model shows low performance even at 1024D, and with heavy hyperparameter tuning. With this large performance gap ($\sim$25\%) between our tree-based and sequential baselines, we conclude that ListOps is an ideal setting to test the parsing ability of latent-tree learning models that are deprived of syntactic supervision.

\paragraph{Latent tree models} Prior work \citep{yogatama2016learning, choi2017unsupervised, maillard2017jointly, williams2018learning} has established that latent tree learning models often outperform standard TreeLSTMs at natural language tasks. In Table~\ref{tab:acc} we summarize results for baseline models and latent tree models on SNLI, a textual entailment corpus. We see that all latent tree models outperform the TreeLSTM baseline, and ST-Gumbel does so with a sizable margin. However, the same models do very poorly on the ListOps dataset. A TreeLSTM model, with its access to ground truth parses, can essentially solve ListOps, achieving an accuracy of 98.7\% with 128D model size. The RL-SPINN and ST-Gumbel models exhibit very poor performance, achieving 64.8\% and 61.0\% accuracy with 128D model size. These latent tree models are designed to learn to parse, and use the generated parses to build sentence representations. Theoretically then, they should be able to find a parsing strategy that enables them to succeed at ListOps. However, their poor performance in this setting indicates that they can not learn a sensible parsing strategy.
%However, their poor performance in this setting where parsing correctly is strongly encouraged, and doing so ensures success, suggests that these models may be incapable of learning a sensible parsing strategy.

Interestingly, the latent tree models perform substantially worse than the LSTM baseline. We theorize that this may be because the latent tree models do not settle on a single parsing strategy. The LSTM can thoroughly optimize given its fully sequential approach. If the latent tree models keep changing their parsing strategy, they will not be able to optimize nearly as well as the LSTM. % TODO: This is argument is weirdly phrased. 

\begin{table}[t!]
\centering
\begin{tabular}{lcccc}
\toprule
 & \multicolumn{2}{c}{\bf Accuracy} & \bf Self  \\
\bf Model & \pmb{$\mu (\sigma)$} & \bf max & \bf F1 \\
\midrule
LSTM & 71.5 (1.5) & \bf 74.4 & - \\
RL-SPINN & 60.7 (2.6) & 64.8 & 30.8 \\
ST-Gumbel & 57.6 (2.9) & 61.0 & \bf 32.3 \\
\midrule
Random Trees & - & - & 30.1\\
\bottomrule
\end{tabular}  
\caption{\label{tab:var} \textit{Accuracy} shows accuracy across four runs of the models (expressed as mean, standard deviation, and maximum). \textit{Self F1} shows how well each of these four model runs agrees in its parsing decisions with the other three.}
\end{table}

To test repeatability and each model's robustness to random initializations, we do four runs of each 128D model (using the best hyperparameter settings); we report the results in Table~\ref{tab:var}. We find that the LSTM maintains the highest accuracy with an average of 71.5. Both latent tree learning models have relatively high standard deviation, indicating that they may be more susceptible to bad initializations.

Ultimately, ListOps is a setting in which parsing correctly is strongly encouraged, and doing so ensures success. The failure of both latent tree models suggests that, in-spite their architectures, they may be incapable of learning to parse.

%%%%%%%%%%%% ANALYSIS %%%%%%%%%%%%
\section{Analysis}

\begin{table}[t!]
\small \centering
\begin{tabular}{lcccc}
\toprule
 & \multicolumn{3}{c}{\bf F1 wrt.} & \bf Avg.  \\
\bf Model & \bf LB & \bf RB & \bf GT & \bf Depth  \\
\midrule
48D~~ RL-SPINN &\bf  64.5 & \bf 16.0 & 32.1 & \bf 14.6 \\
128D RL-SPINN & 43.5 & 13.0 & \bf 71.1 & 10.4 \\
%\midrule
48D~~ ST-Gumbel & 52.2 & 15.3 & 55.3 & 11.1 \\
128D ST-Gumbel & 56.5 & 9.8 & 57.3 & 12.7 \\
\midrule
Ground-Truth Trees & 41.6 & 8.8 & 100.0 & 9.6 \\
Random Trees & 24.0 & 24.0 & 24.2 & 5.2 \\
\bottomrule
\end{tabular}  
\caption{\label{tab:f1} \textit{F1 wrt.} shows F1 scores on ListOps with respect to left-branching (LB), right-branching (RB), and ground-truth (GT) trees. \textit{Avg. Depth} shows the average across sentences of the average depth of each token in its tree.}
\end{table}

Given that the latent tree models perform poorly on ListOps, we take a look at what kinds of parses these models produce. 

\paragraph{F1 scores} In Table~\ref{tab:f1}, we show the F1 scores between each model's predicted parses and fully left-branching, right-branching, and ground-truth trees. We use the best run for each model in the reported statistics.

Overall, the RL-SPINN model produces parses that are most consistent with the ground-truth trees. The ListOps ground-truth trees have a high F1 of 41.6 with left-branching trees, compared to 9.8 with right-branching trees. \citet{williams2018learning} show that RL-SPINN tends to settle on a left-branching strategy when trained on MultiNLI. We observe a similar phenomena here at 48D. Since ListOps is more left-branching, this tendency of RL-SPINN's could offer it an advantage. Furthermore, as might be expected, increasing model size from 48D to 128D helps improve RL-SPINN's parsing quality. At 128D, it has a high F1 score of 71.1 with ground-truth trees. The 128D model also produces parses with an average tree depth (10.4) closer to that of ground-truth trees (9.6). 

The parses from the 128D ST-Gumbel have a significantly lower F1 score with ground-truth trees than the parses from RL-SPINN. This result corresponds with the performance on the ListOps task where RL-SPINN outperforms ST-Gumbel by $\sim$4\%. Even though the trees ST-Gumbel generates are of a worse quality than RL-SPINN's, the trees are consistently better than random trees on F1 with ground-truth trees. %Clearly then, the models are learning something non-trivial about parsing in ListOps 

It's important to note that the F1 scores have very high variance from one run to the next. Table~\ref{tab:var} shows the self F1 scores across random restarts of both models. Both have very poor agreement in parsing decisions across restarts, their self F1 is comparable to that of randomly generated trees. For RL-SPINN, the F1 with ground-truth trees ranges from 18.5 to 71.1, with an average of 39.8 and standard deviation of 19.4. While ST-Gumbel has an average of 44.5, and a standard deviation of 11.8. This high variance in F1 scores is reflective of the high variance in accuracy across random restarts, and it supports our hypothesis that these latent tree models do not find and settle on a single parsing strategy.

\paragraph{Parse trees} In Figure~\ref{fig:trees}, we show some examples of trees generated by both models. We use the best runs for the 128D versions of the models. Parses generated by RL-SPINN are in the left column, and those generated by ST-Gumbel are on the right.

For the pair of examples in the top row of Figure~\ref{fig:trees}, both models incorrectly predict 5 as the solution. Both parses compose the first three operations together, and it is not clear how these models arrive at that solutions given their chosen parses. 

In the second pair of examples, RL-SPINN predicts the correct value of 7, while ST-Gumbel wrongly predicts 2. The parse generated by RL-SPINN is not the same as the ground-truth tree but it finds some of the correct constituent boundaries: $\big([\MAX~5~6\big)$ are composed with a right-branching tree, and $\big(2~]\big)$ are composed together. Since the first three operations are all \textsc{sum\_mod}, their strange composition does not prevent the model from correctly predicting 7.

For the third pair of examples, the RL-SPINN model generates the same parse as the ground-truth reference and rightly predicts 6. While ST-Gumbel gets some of the correct constituent boundaries, it produces a fairly balanced tree, and falters by predicting 5. Overall, the generated parses are not always interpretable, particularly when the model composes several operations together. 

%However, when a model generates the correct parse, or something close, it is capable of predicting the correct value. This shows that the difficulty lies in parsing ListOps, and the failure of the models lies in their failure to properly parse. % TODO: Give some numbers or cut this---hard to know what capable means otherwise.
\iffalse
We find that when RLSPINN generates the reference parse, it correctly classifies the example 98\% of the time (96\% for ST-Gumbel). About half of these examples, where the generated parse exactly matched the reference parse, are shorter sequence ($<23$ tokens). 70\% of these shorter examples are classified correctly by RL-SPINN. However, when the parse is correct, the prediciton accuracy is 98\%.
\fi

\begin{figure}[t!]
  \centering
    \includegraphics[width=0.5\textwidth]{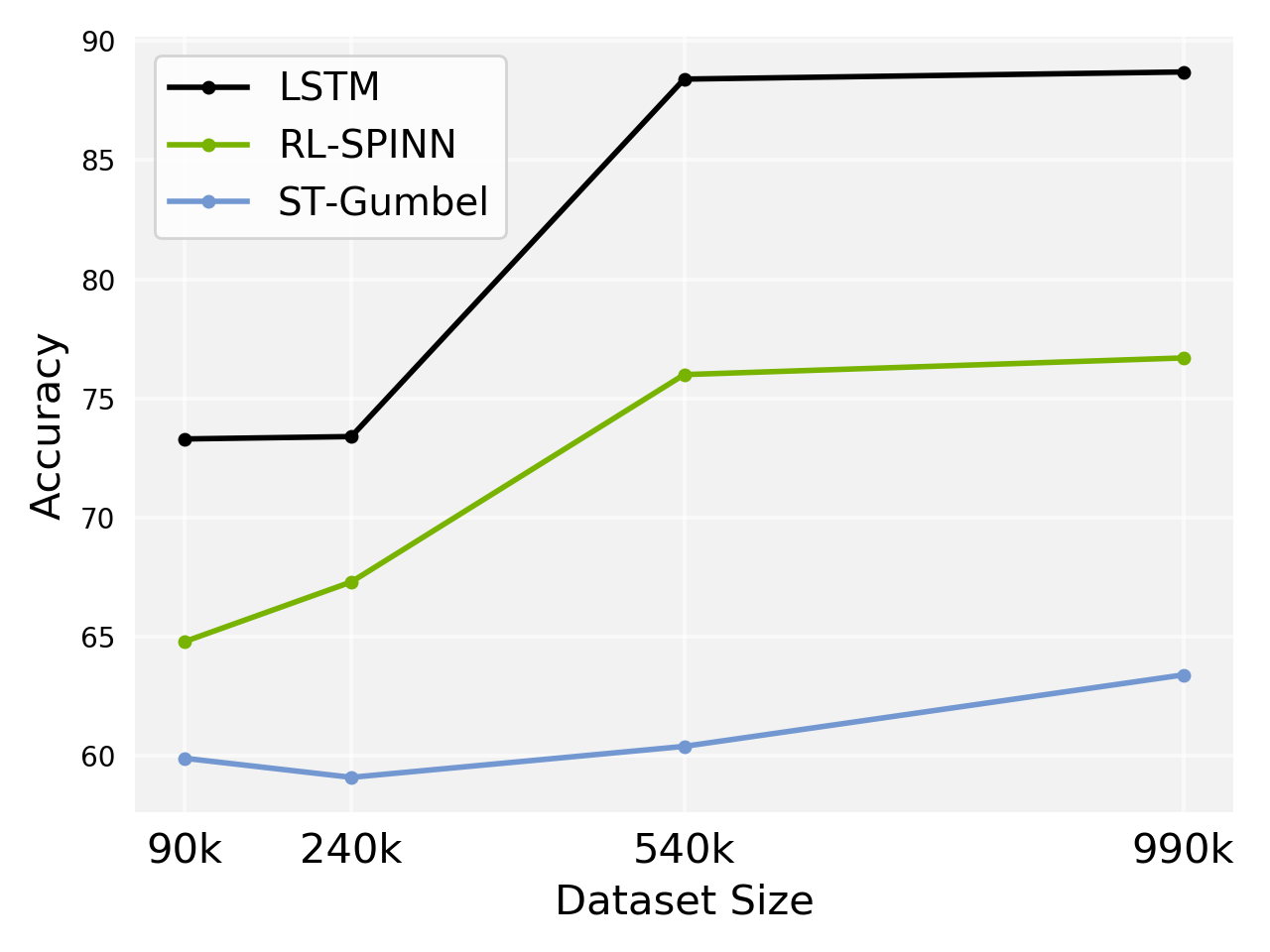}
   	\caption{Model accuracy on ListOps test set by size of training dataset.}
	\label{fig:datasize}
\end{figure}

\paragraph{Dataset size} ListOps is intended to be a simple dataset that can be easily solved with the correct parsing strategy. One constraint on ListOps is the dataset size. With a large enough dataset, in principle an RNN with enough capacity should be able to solve ListOps. As we stated in Section~3, a requirement for ListOps is having a large RNN--TreeRNN gap to ensure the efficacy of the dataset.

However, it is possible that the latent tree models we discuss in this paper could greatly benefit from a larger dataset size, and may indeed be able to learn to parse given more data. To test this hypothesis, and to ensure that data volume is not critical to solving ListOps, we generate three expansions on the training data, keeping the original test set. The new training datasets have 240k, 540k, and 990k examples, with each dataset being a subset of the next larger one. We train and tune the 128D LSTM, RL-SPINN, and ST-Gumbel models on these datasets. Model accuracies for all training sets are plotted in Figure~\ref{fig:datasize}. We see that while accuracy does go up for the latent tree models, it's not at a rate comparable to the LSTM. Even with an order of magnitude more data, the two models are unable to learn how to parse successfully, and remain thoroughly outstripped by the LSTM. Clearly then, data volume is not a critical issue keeping these latent tree models from success.

% TODO: Comment on RLSPINN benefiting more greatly from larger training data that ST-Gumbel?

\section{Conclusion}
In this paper we introduce ListOps, a new toy dataset that can be used as a diagnostic tool to study the parsing ability of latent tree learning models. ListOps is an ideal setting for testing a system's parsing ability since it is explicitly designed to have a large RNN--TreeRNN performance gap. While ListOps may not be the simplest type of dataset to test a system's parsing capability, it is certainly simpler than natural language, and it fits our criteria. 

The experiments conducted on ListOps with leading latent tree learning models show that these models are unable to learn to parse, even in a setting that strongly encourages it. We only test two latent tree models, and are unable to train and analyse some other leading models, like \citeauthor{maillard2017jointly}'s (\citeyear{maillard2017jointly}) due to its high computational complexity. In the future, we would like to develop a version of ListOps with shorter sequence lengths, while maintaining the RNN--TreeRNN gap. With such a version, we can experiment with more computationally intensive models.

Ultimately, we aim to develop a latent tree learning model that is able to succeed at ListOps. If the model can succeed in this setting, then perhaps it will discover interesting grammars in natural language that differ from expert designed grammars. If those discovered grammars are principled and systematic, they may lead to improved sentence representations. We hope that this work will inspire more research on latent tree learning and lead to rigorous testing of such models' parsing abilities.

\section*{Acknowledgments}

This project has benefited from financial support to Sam Bowman by Google, Tencent Holdings, and Samsung Research. We thank Andrew Drozdov, who contributed to early discussions that motivated this work. 

\bibliographystyle{acl_natbib} 
\bibliography{naacl}

\end{document}